# STUDY OF A ROBUST ALGORITHM APPLIED IN THE OPTIMAL POSITION TUNING FOR THE CAMERA LENS IN AUTOMATED VISUAL INSPECTION SYSTEMS

**Radu Arsinte**
Software ITC S.A.- Str Republicii 109
3400 Cluj-Napoca
Romania
Tel.+40-64-197681  Fax.+40-64-196787
email:radu@sitc1.dntcj.ro

*Abstract*
*This paper present the mathematical fundaments and experimental study of an algorithm used to find the optimal position for the camera lens to obtain a maximum of details. This information can be further applied to a appropriate system to automatically correct this position .The algorithm is based on the evaluation of a so call "resolution function" who calculates the maximum of gradient in a certain zone of the image. The paper also presents alternative forms of the function , results of measurements and set up a set of practical rules for the right application of the algorithm.*

## 1. Introduction

In many applications in visual industrial inspection systems [4] , is needed to evaluate the image resolution , or to be more specific , maximum of the gradient function for luminance in the specific area. The first and maybe the most important application is optimum adjustment of lens position in image acquisition systems , to obtain maximum of details in acquisition process. The proposed method is based on the considerations found in [1].

## 2. Description of the proposed method

For simplicity we suppose that the original image is a plane image. The image generated from the optical-electrical system is affected of the linear transform in the following formula:

$$U(x,y) = \int\int_{-\infty}^{+\infty} V(x-\xi, y-\eta) h(\xi, \eta) d\xi d\eta \qquad (1)$$

In this formula h is the transient transfer function of the linear system. For an optimal position (optimal optical focusing) this function is done by the aperture characteristic of the videocaption system. In the case of the large defocusing, done by a displacement of the lens relative to the photocathode equal to z, this function is delimited in a distributed surface of a radius R given by a punctual source, like in the following formula:

$$h(x, y) = \begin{cases} \dfrac{1}{\pi R^2} & \text{for } x^2 + y^2 < R^2 \\ 0 & \text{all other cases} \end{cases} \quad (2)$$

where the radius R is done by the formula:

$$R = \frac{A - F}{2AG} z \quad (3)$$

In the precedent formula:

A-distance to the object, F-focal length, G-light intensity

We suppose that the in the original image we have a sharp transition in the luminance function:

$$V(x, y) = \begin{cases} 1 & \text{for } x \geq 0 \\ 0 & \text{in all other cases} \end{cases} \quad (4)$$

Therefore, for y=0 in (1) we have:

$$U(x) = \int_{\infty}^{x} h_x(\xi) d\xi \quad (5)$$

where:

$$h_x(\xi) = \int_{-\infty}^{+\infty} h(\xi, \eta) d\eta \quad (6)$$

Developing the input function on the x coordinate: $\dfrac{dU}{dx} = h_x(x)$

For the transient function given in (6) we have :

$$\frac{dU}{dx} = \frac{2}{\pi R^2} \sqrt{R^2 - x^2} \quad (7)$$

The maximum of the function is done for x=0. Using the precedent formulae and considering the simetry, we can write the final formula reflecting the content of details in the image (resolution function):

$$D(z) = \max_x \left( \frac{dU}{dx} \right) = \frac{4AG}{\pi(A - F)|z|} \quad (8)$$

Formula (8) reflects the resolution for large values of |z| , in a case a large defocusing. For small values of z , the resolution is limited to a value $D_{max}$ :

$$D_{max} = \max_{x}\left(\int_{-\infty}^{+\infty} h(x,y)dy\right) \tag{9}$$

where h is the aperture characteristic for the videocapture device

The plot of simulation for the precedent formula (8) is presented in Figure 1.

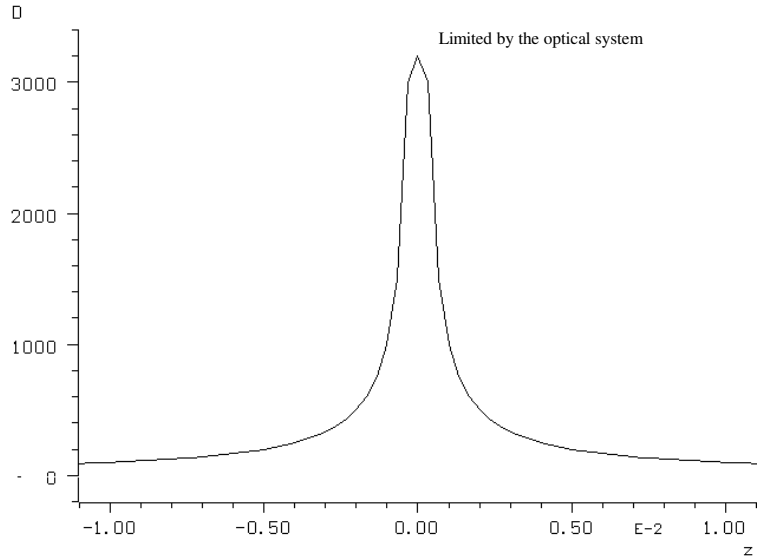

**Fig.1. Plot of the resolution function**

### 3. Experimental results

In the following measurements presented, the discrete variant of formula used for the resolution function is :

$$d = \sum_{i=1}^{N-1}\sum_{j=1}^{N-1}\{(e_{i,j}-e_{i+1,j+1})^2 + (e_{i+1,j}-e_{i,j+1})^2\} \tag{10}$$

Where:

$e_{i,j}$ is the sample corresponding to (i,j) coordinates in the calculation window (nucleus).

d- resolution function    N-nucleus dimension

Other practical form of the resolution function which can be used:

$$d = \sum_{i=1}^{N-1}\sum_{j=1}^{N-1} |e_{i,j}-e_{i+1,j+1}| + |e_{i+1,j}-e_{i,j+1}| \tag{11}$$

The two formulae (10) and (11) give similar results. The choice is done by the simplicity of the computing process. The author has chosen the formula (10) applying the algorithm in a DSP based system, where the multiplication operation is done in a single cycle, therefore formula (10) is easy to program and is computed faster than (11).

Practical implementation of the algorithm was done initially on 7X7 pixels nucleus. The function returns a numerical parameter related with the amount of details in the investigated window. Experiments has revealed that in real conditions (noise added in the image) the stability of determination is seriously affected. Therefore the nucleus was raised up to 31x31 pixels , in which case the measurements are more stable, as a result of the averaging effect for the noise. The image processing simulation environment , described in [3], allows us the possibility to freely choose the window dimension.

Effect of window dimension on the stability of computing for the resolution function is illustrated in the following table (Table 1).We can observe a better stability of measurements done with a large dimension of nucleus. For large values of the nucleus the limit of accuracy seems to be only the metrologic proprieties of the acquisition system [2].

**Table 1. Results for various nucleus dimensions**

| Nucleus dimension (pixels) | Resolution (3 measurements) | Average value | Dispersion |
|---|---|---|---|
| 5x5 | 381 | 346 | +10% |
| | 290 | | -16% |
| | 366 | | +5,7% |
| 9x9 | 1368 | 1377 | -0,69% |
| | 1352 | | -1,8% |
| | 1412 | | +2,5% |
| 17x17 | 12548 | 12594 | -0,36% |
| | 12479 | | -0,91% |
| | 12756 | | +1,27% |
| 31x31 | 27930 | 27898 | 0,11% |
| | 27896 | | -0,007% |
| | 27868 | | -0,1% |

Following images (Figure 2) present the practical use of the resolution function in the optimal adjustment of optical focusing system in visual inspection systems.

The second image is the optimal focused image (maximum resolution) ,the first is the same image defocused. Evaluations are done around the same point of the image. We can see a considerable large value of the resolution function in the second case (optimal focusing).

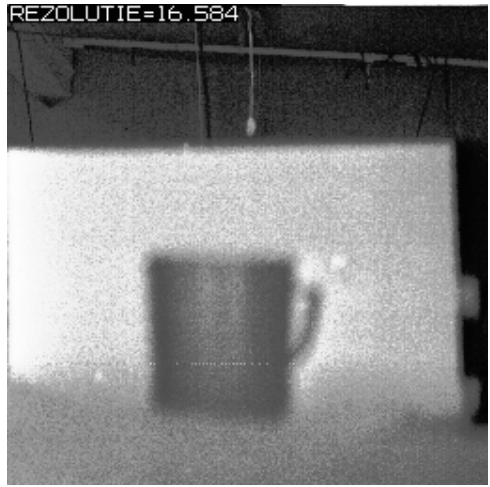 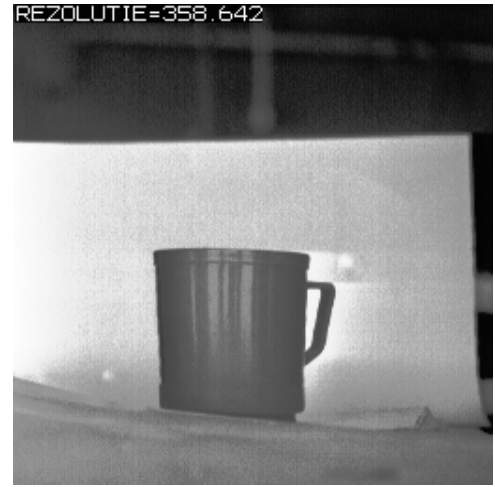

          (a)  R=16584                                     (b) R=358642

**Fig.2. Values of the resolution function for a defocused (a) and optimal (b) image of the same object**

### 4. Conclusion

The presented algorithm seems to be a reasonable solution for the problem of optimal focusing in automated visual inspection systems [4]. The complexity of algorithm is low resulting in fast calculations and the measurements , in given conditions, are stable. Future enhancements are related to an optimal finding of the reference window to make the measurements insensitive to the choice of position. Similar algorithms are used in modern digital cameras.